\RequirePackage{fix-cm}
\documentclass[smallextended]{svjour3}       % onecolumn (second format)
\smartqed  % flush right qed marks, e.g. at end of proof

\PassOptionsToPackage{numbers,compress}{natbib}
\usepackage[utf8]{inputenc} % allow utf-8 input
\usepackage[T1]{fontenc}    % use 8-bit T1 fonts

\usepackage{url}            % simple URL typesettinghttps://www.overleaf.com/project/603200ac3f7784096d301a16
\usepackage{booktabs}       % professional-quality tables
\usepackage{amsfonts}       % blackboard math symbols
\usepackage{nicefrac}       % compact symbols for 1/2, etc.
\usepackage{microtype}      % microtypography
\usepackage[sectionbib]{natbib}

\usepackage{multirow}
\usepackage{textcomp}

\usepackage{graphicx}
\usepackage{caption}
\usepackage{subcaption}

\usepackage{float}

\usepackage{algorithm}
\usepackage{algpseudocode}
\usepackage[colorinlistoftodos]{todonotes}

\usepackage{amsmath,amssymb,url,longtable,booktabs,multicol,multirow,lineno}

\usepackage[multiple]{footmisc}
\usepackage{tikz}
\usepackage{tikzscale}
\usepackage{xspace}
\usepackage{pgfplots}
\usepackage{todonotes}
\usepackage{comment}

\usepackage{marvosym}

\usepackage[T1]{fontenc}
\usepackage{lmodern}

\DeclareUnicodeCharacter{20AC}{\EUR{}}
\usepgfplotslibrary{groupplots}
\pgfplotsset{compat=1.16}
\nolinenumbers

\usepackage{hyperref}       % hyperlinks

\begin{document}
%----------------------------------------------------------------------

\title{Efficient Analysis of COVID-19 Clinical Data using Machine
  Learning Models}


\author{Sarwan Ali \and Yijing Zhou \and Murray Patterson}


\institute{Sarwan Ali \at
  Georgia State University, Atlanta, Georgia, USA \\
  \email{sali85@student.gsu.edu}
  \and
  Yijing Zhou \at
  Georgia State University, Atlanta, Georgia, USA \\
  \email{yzhou43@student.gsu.edu}
  \and 
  Murray Patterson \at
  Georgia State University, Atlanta, Georgia, USA \\
  \email{mpatterson30@gsu.edu}
}

\date{Received: date / Accepted: date}

\maketitle

\begin{abstract}

Because of the rapid spread of COVID-19 to almost every part of the globe, huge volumes of data and case studies have been made available, providing researchers with a unique opportunity to find trends and make discoveries like never before, by leveraging such big data.  This data is of many different varieties, and can be of different levels of veracity e.g., precise, imprecise, uncertain, and missing, making it challenging to extract important information from such data.  Yet, efficient analyses of this continuously growing and evolving COVID-19 data is crucial to inform --- often in real-time --- the relevant measures needed for controlling, mitigating, and ultimately avoiding viral spread.  Applying machine learning based algorithms to this big data is a natural approach to take to this aim, since they can quickly scale to such data, and extract the relevant information in the presence of variety and different levels of veracity.  This is important for COVID-19, and for potential future pandemics in general.

In this paper, we design a straightforward encoding of clinical data (on categorical attributes) into a fixed-length feature vector representation, and then propose a model that first performs efficient feature selection from such representation.  We apply this approach on two clinical datasets of the COVID-19 patients and then apply different machine learning algorithms downstream for classification purposes. We show that with the efficient feature selection algorithm, we can achieve a prediction accuracy of more than 90\% in most cases. We also computed the importance of different attributes in the dataset using information gain.  This can help the policy makers to focus on only certain attributes for the purposes of studying this disease rather than focusing on multiple random factors that may not be very informative to patient outcomes.

  
\end{abstract}

\section{Introduction}
%----------------------------------------------------------------------

Because of the rapid global spread of COVID-19, and the cooperation of
medical institutions worldwide, a tremendous amount of public data ---
more data than ever before for a single virus --- has been made
available for
researchers~\cite{gisaid_website_url,ali2021spike2vec,ali2021classifying}.
This ``big data'' opens up new opportunities to analyse the behavior
of this virus~\cite{leung2020big,leung2020machine}.  Despite these
opportunities, the huge size of the data poses a challenge for its
processing on smaller systems~\cite{ali2021spike2vec}.  On one hand,
this creates scalability issues, and on the other hand, it creates the
problem of high dimensionality (the curse of
dimensionality)~\cite{Ali2020ShortTerm,ali2019short}.  Since such data
was not previously available to the research community at this
magnitude and ease of access, new and more sophisticated methods are
required to extract useful information from this big data.

At the same time, the shortage of medical resources may occur when
such a severe pandemic happens, and the surging number of patients
exceeds the capacity of the clinical system.  This situation happens
in many countries and regions during continuous outbreaks of the
COVID-19 pandemic, and clinicians have to make the tough decision of
which individual patients have a higher possibility to recover and
should receive a limited amount of medical care.  What is more
difficult is the decision of which patients have little to no chance
of survival, regardless of treatment level, and should hence be
abandoned for the sake of optimizing the use of limited resources for
others who still have a chance.  In addition to this, patients with
different levels of severity and urgency of symptoms require the
medical system to create a complete plan to provide various treatments
in proper order~\cite{abdulkareem2021realizing}.

Hence, a clinical decision support system is of utmost importance to
optimize the use of the limited medical resources and thus save more
lives overall~\cite{abdulkareem2021realizing}.  In order to develop a such a clinical decision support
system with high quality, it is necessary to build a model that can
predict the possible complications of patients, assessing the
likelihood that they will survive under certain levels of care.
Machine learning (ML) based algorithms are proven to perform well in
terms of classification and clustering.  Therefore, we work on
building machine learning models that can scale to larger datasets and
reduce the run time by selecting the proper attributes.  Since ML
models take a feature vector representation as input~\cite{ali2021predicting,grover2016node2vec}, designing such
vectors while preserving a maximum amount of information is a
challenging task~\cite{yang2018multi}.  Moreover, when the size of data becomes this large,
even scalability of these models becomes an issue.  Much work has been
done on genomic data of COVID-19 patients~\cite{kuzmin2020machine,ali2021effective,melnyk2021alpha}.  One major challenge in
this case is the conversion of genomic sequences into fixed-length
numerical feature vectors so that they can be given as input to the
underlying ML classifiers for prediction purposes.

%Many studies about COVID-19 clinical have been conducted to discover
%more information on coronavirus and help medical doctors provide
%better treatments for patients. However, the considerable size of the
%COVID-19 dataset may cause trouble for researchers.\todo{Murray: few
%citations here?  Maybe from related work section?} It's hard to
%handle oversized data and solve the scalability issues, and the
%program run time will also increase significantly with the growth of
%data size.

In this paper, we propose an algorithm that efficiently predicts with
high accuracy patient mortality as a function many different factors.
This problem can help doctors to prescribe medications and design
strategies in advance that can help to save the highest number of
lives.  In this paper, our contributions can be summarised as follows:
\begin{enumerate}
\item We propose a pipeline to efficiently predict patient mortality
  as a function of a few dozen factors.  We show that with basic
  information about a patient (gender, race, exposure, etc.), we can
  predict in advance the likelyhood of a mortality in the future.  We
  also predict if a patient is COVID-19 positive or negative using
  attributes like red blood cells and hemoglobin, etc.
\item We show that our model is scalable on larger dataset (achieves
  accuracies $>$90\%).
\item From our results, it is evident that the proposed model (using
  efficient feature selection) outperforms the baselines (without
  using any feature selection) in terms of prediction accuracy and
  runtime.
\item We show the importance of each attribute by measuring the
  information gain of the attributes with the class labels.  This
  study can help doctors and other relevant authorities to focus more
  on specific attributes rather than dealing with all information at
  once, which can be difficult for humans to fathom.
\end{enumerate}

The rest of the paper is organised as follows:
Section~\ref{sec_related_work} contains literature review for the
problem. Our proposed model is give in
Section~\ref{sec_proposed_approach}. Dataset statistics and
experimental details are given in
Section~\ref{sec_experimental_setup}. We show results and their
comparisons in Section~\ref{sec_results}. We discuss the importance of
different attributes in
Section~\ref{sec_attribute_importance}. Finally, in
Section~\ref{sec_conclusion}, we conclude our paper.

\section{Related Work}\label{sec_related_work}
% \textcolor{blue}{
Machine learning based models that take fixed length feature vectors as input has been successfully applied (for data analytics tasks) in many domains such as graphs analytics~\cite{ali2021predicting,AHMAD2020Combinatorial}, smart grid~\cite{ali2019short,Ali2020ShortTerm}, electromyography (EMG)~\cite{ullah2020effect}, and text classification~\cite{Shakeel2020LanguageIndependent,Shakeel2020Multi,Shakeel2019MultiBilingual}. It is important to perform objective evaluation of the underlying model rather than just doing subjective evaluation~\cite{hassan2021locally}. 
Many methodologies of data science have been applied to objectively analyze the data of COVID-19 and provide support to the medical system. The synergy between data scientists and the biomedical communities is helpful to improve the detection of diseases and illnesses, as well as the prediction of possible complications. 
% For instance, previous researchers have 
Authors in~\cite{leung2020data} developed a framework with to predict cancer trends accurately. This type of framework could be used for the analysis of other clinical data.
% knowledge about the desease from healthcare data by using a data mining approach.
S. Ali et al. in~\cite{ali2021k} uses spike sequences to classify the variants of the COVID-19 infected humans. An effective approach to cluster the spike sequences based on the virus's variants is conducted in~\cite{ali2021effective}.

Several studies discussed different data mining techniques to study the behavior of the COVID-19~\cite{li2020using,albahri2020role,leung2020big}.
% Authors in~\cite{li2020using} do systematic reviews about the role of data science techniques, including Machine Learning and Data Mining, in medical research on COVID-19. 
% Authors in~\cite{li2020using} demonstrate that computational methods trained on large clinical datasets can generate more reliable COVID-19 diagnostic models to mitigate the medical pressure of cases testing[3]. 
Authors in~\cite{fung2021predictive} uses neural networks, which takes advantage of few-shot learning and autoencoder to perform  predictive analysis on COVID-19 data.
% Data science solutions for COVID-19 predictive analysis can be achieved by using Neural networks (NNs), which takes advantage of few-shot learning and autoencoder and allows the model to get trained on small samples of more accessible and less expensive kinds of data like antibody test results from blood samples~\cite{fung2021predictive}. 
% One practice of big data science on the COVID-19 dataset is to conduct effective analysis on some attributes, including age and genders, and enable the users to decide whether or not to ignore the Null/unstated data~\cite{leung2020big}. 
% Some existing work on COVID-19 data focuses on 
% reporting the spatial differences and temporal trends of the numbers of confirmed cases and mortality, while some work on exploring the variance among different gender and age groups instead. By examing the data of confirmed cases and mortality on the < gender, age group > combinations, the algorithm finds out the most vulnerable gender and age groups of COVID-19, which is helpful for clinicians to provide preventive approaches. 
A study for predicting the clinical outcomes of patients and indicating whether patients are more likely to recover from coronavirus or in danger of death is performed in~\cite{leung2020machine}. They presented a tool called online analytical processing (OLAP), which can help the researchers learn more about the confirmed cases and mortality of COVID-19 by conducting machine learning methods on the big dataset of COVID-19. 

\section{Proposed Approach}\label{sec_proposed_approach}
Most of the machine learning (ML) models take fixed-length feature vectors as an input to perform different tasks such as classification and clustering. 
We design a fixed-length feature vector representation, which includes the values of different attributes of the clinical data. 
One important point to mention here is that not all the features in the vectors are important in terms of predicting the class labels. Therefore, it is required to apply feature selection to not only improve the predictive performance of the underlying classifiers (by removing unnecessary features), but also improve the training runtime. 
The feature selection methods that we used in this paper are discussed below.
% \todo{Murray: it is not clear what the feature vector representation is}

\subsection{Feature Selection Methods}
We use different supervised and unsupervised feature selection methods to improve the runtime of the underlying classifiers and also to improve the predictive performance. For supervised models, we use Boruta (shadow features)~\cite{kursa2010feature}, and Ridge Regression (RR)~\cite{hoerl1975ridge}. For unsupervised methods, we use Approximate kernel approach called Random Fourier Features (RFF)~\cite{rahimi2007random}.
% https://towardsdatascience.com/boruta-explained-the-way-i-wish-someone-explained-it-to-me-4489d70e154a
\subsubsection{Boruta (shadow features)}
The main idea of Boruta is that features do not compete among themselves but rather they compete with a randomized version of them. 
Boruta captures the non-linear relationships and interactions using the random forest algorithm. It then extract the importance of each feature (corresponding to the class label) and only keep the features that are above a specific threshold of importance. To compute the importance of the features, it performs the following task: From the original features set in the data, it creates dummy features (shadow features) by randomly shuffling each feature. Now the shadow features are combined with the original features set to obtain a new dataset, which has twice the number of features of the original data. Using random forest, it compute the importance of the original features and the shadow features separately. Now the importance of the original features are compared with the threshold. The threshold is defined as the highest feature importance recorded among the shadow features. The feature from the original feature set is selected is its importance (computed using random forest) is greater than the threshold (highest importance value among shadow features). In Boruta, a feature is useful only if it is capable of doing better than the best randomized feature. Note that we are using two datasets in this paper, namely Clinical Data1, and Clinical Data2 (see Section~\ref{sec_dataset_statistics} for detail regarding datasets).
For the Clinical Data1, Boruta selected $11$ features out of $19$ and removed Year, Gender, Race, Case Positive Specimen Interval, Case Onset Interval, Exposure, Current Status, and Symptom Status.
For the Clinical Data2, Boruta selected $7$ features from $18$ features in total. The selected features are  Red blood Cells, Platelets, Hematocrit, Monocytes, Leukocytes, Eosinophils, and Proteina C reativa mg/dL.

\subsubsection{Ridge Regression}
Ridge Regression (RR) is a supervised algorithm for parameter estimation that is used to address the collinearity problem that arises in multiple linear regression frequently \cite{mcdonald2009ridge}.
Its main idea is to increase the bias (it first introduces a Bias term for the data) to improve the variance, which shows the generalization capability of RR as compared to simple linear regression. RR ignores the datapoints that are far away from others and it try to make the regression line more horizontal. RR is useful for Feature selection because it gives insights on which independent variables are not very important (can reduce the slope close to zero). The un-important  independent variables are then removed to reduce the dimensions of the overall dataset. The objective function of ridge regression is the following
\begin{equation}
    min(\text{Sum of square residuals} + \alpha \times \text{slope}^2)
\end{equation}
where $\alpha \times {slope}^2$ is called penalty terms.

\subsubsection{Random Fourier Features (RFF)}
% When dealing with ``Big Data", typical machine learning (ML) algorithms does not scale directly~\cite{ali2021spike2vec}. This requires some sort of preprocessing of the data initially so that dimensionality of the data can be reduced.
A popular approach for the classification is using kernel based
algorithms, which computes a similarity matrix that can be used as input for traditional classification algorithms such as support vector machines. However, pair-wise computation for the kernel matrix is an expensive task. To make this task efficient, an efficient approach, called kernel trick is used.
\begin{definition}[Kernel Trick]
It works by taking dot product between the pairs of input points.
%   It is used to generate features for an algorithm which  depends on the inner product between the pairs of input data  points. 
  Kernel trick avoid the need to map the input data
  (explicitly) to a high-dimensional feature space.
\end{definition}
The main idea of the Kernel Trick is the following: \textit{Any
  positive definite function f(x,y), where $x,y \in \mathcal{R}^d$,
  defines an inner product and a lifting $\phi$ for the purpose of computing the inner product quickly between the lifted data points}~\cite{rahimi2007random}. More formally:
\begin{equation}
  \langle \phi (x), \phi (y) \rangle = f(x,y)
\end{equation}
The main problem of kernel method is that when we have large sized data, they suffers from large initial computational and storage costs.
To solve these problems, we use an approximate kernel method called Random Fourier Features (RFF)~\cite{rahimi2007random}. The RFF maps the given data to a low dimensional randomized feature space (euclidean inner product
space). More formally:
\begin{equation}\label{eq_rff_1}
  z: \mathcal{R}^d \rightarrow \mathcal{R}^D
\end{equation}
RFF basically approximate the inner product between a pair of
transformed points. More formally:
\begin{equation}\label{eq_z_value}
  f(x,y) = \langle \phi (x), \phi (y) \rangle \approx z(x)' z(y)
\end{equation}
In Equation~\eqref{eq_z_value}, $z$ is low dimensional (unlike the
lifting $\phi$). In this way, we can transform the original input data
with $z$. Now, $z$ is the approximate low dimensional embedding for
the original data. We can then use $z$ as the input for different classification algorithms.

\subsection{Classification Algorithms}
For classification, we use Support Vector Machine (SVM), Naive Bayes (NB), Multiple Linear Regression (MLP), K-Nearest Neighbors (KNN), Random Forest (RF), Logistic Regression (LR), and Decision Tree (DT). All algorithms are used with default parameters. The value for K in case of KNN is taken as $5$ (using standard validation set approach~\cite{validationSetApproach}).

We are also using a deep learning model, called Keras Classifier for the classification purposes. For this model, we use a sequential constructor. We create a fully connected network with one hidden layer that contains $p$ neurons, where $p$ is equal to the length of the feature vector. We use "rectifier" as an activation function and ``softmax" activation function in the output layer. We also use ab efficient Adam gradient descent optimization algorithm with ``sparse categorical crossentropy" loss function (used for multi-class classification problem), which computes the crossentropy loss between the labels and predictions. The batch size and number of epochs are taken as $100$ and $10$, respectively for training the model. Since deep learning model does not require any feature selection, we use the original data without any feature selection as input to keras classifiers.
\begin{remark}
  We use ``sparse categorical crossentropy" instead of simple ``categorical crossentropy" because we are using integer labels rather than one-hot representation of labels.
\end{remark}

\section{Experimental Setup}\label{sec_experimental_setup}
In this section, we describe our dataset in detail. All experiments are performed on a Core i5 system running the Windows operating system, 32GB memory, and a 2.4 GHz
processor.  Implementation of the algorithms is done in Python. Our code and prepossessed dataset is available
online\footnote{\url{https://github.com/sarwanpasha/COVID_19_Clinical_Data_Analytics}}.

\subsection{Dataset Statistics}\label{sec_dataset_statistics}
In this paper, we are using clinical data from two different sources. The description of both datasets is given below.

\subsection{Clinical Data1}
\begin{figure}[h!]
    \centering
    \includegraphics[scale = 0.6]{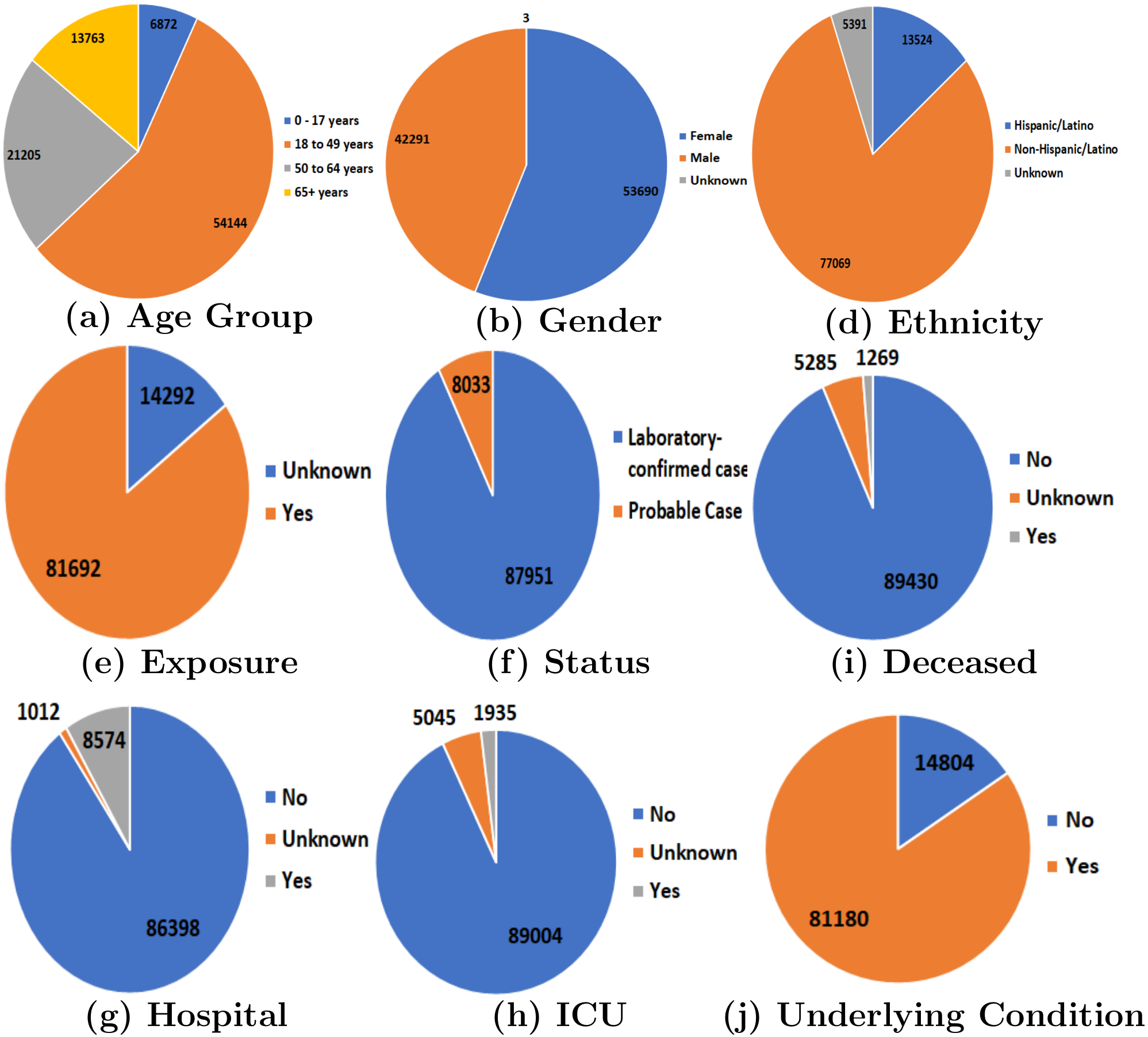}
    \caption{Pie charts for the distribution of different attributes values}
    \label{fig_all_pie_charts}
\end{figure}

\begin{figure}[h!]
    \centering
    \includegraphics[scale = 0.58]{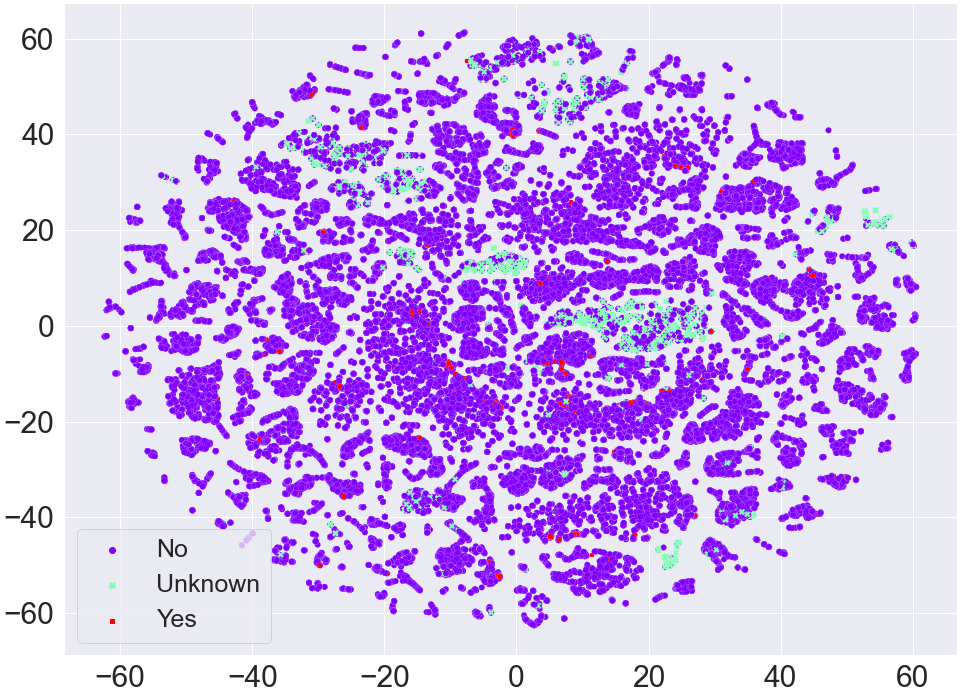}
    \caption{t-SNE plot for deceased attribute of Clinical Data1.}
    \label{fig_t_sne}
\end{figure}

We use COVID-19 Case Surveillance dataset (we call it Clinical Data1 for reference), which is publicly available on center of disease control CDC, USA's website\footnote{\url{https://data.cdc.gov/Case-Surveillance/COVID-19-Case-Surveillance-Public-Use-Data-with-Ge/n8mc-b4w4/data}}. After preprocessing (removing missing values), we got $95984$ patients data record. The attributes in the dataset are following: 

\begin{enumerate}
    \item \textbf{Year:} The earlier of year the Clinical Date. date related to the illness or specimen collection or the Date Received by CDC
    \item \textbf{Month:} The earlier of month the Clinical Date. date related to the illness or specimen collection or the Date Received by CDC (see Figure~\ref{fig_month_year} for month and year distribution).
    \begin{figure}[h!]
        \centering
        \includegraphics[scale = 0.6]{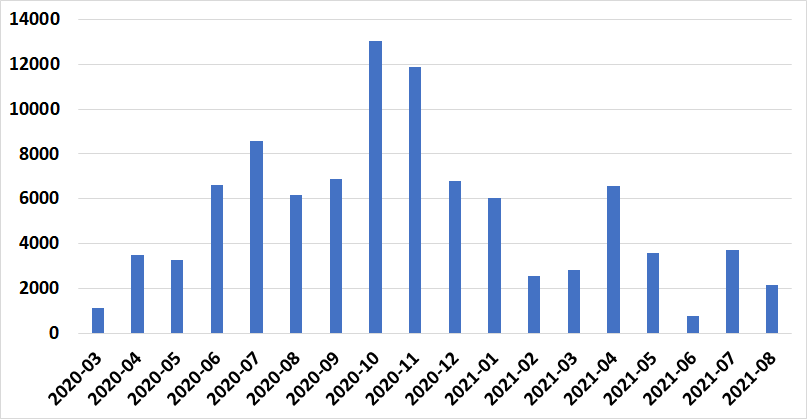}
        \caption{Month and Year attribute distribution.}
        \label{fig_month_year}
    \end{figure}
    \item \textbf{State of residence:} This attributes shows the name of the state (of United States of America) in which the patient is living (see Figure~\ref{fig_states_distribution} for states distribution).
    \begin{figure}[h!]
        \centering
        \includegraphics[scale = 0.6]{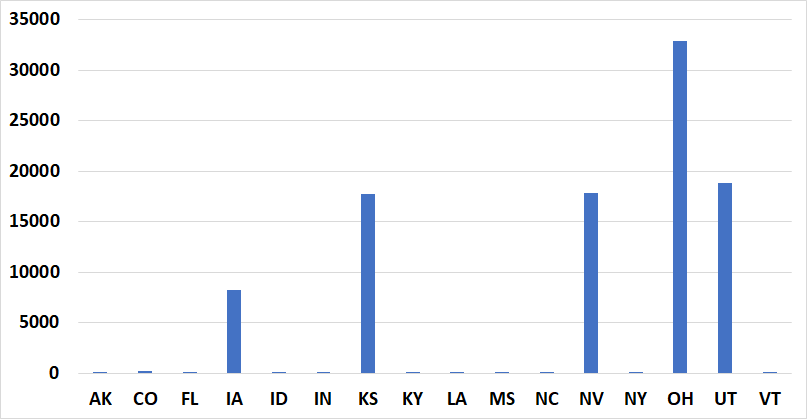}
        \caption{State of residence distribution.}
        \label{fig_states_distribution}
    \end{figure}
    \item \textbf{State FIPS code:} Federal Information Processing Standards (FIPS) code for different states
    \item \textbf{County of residence:} Name of the County
    \item \textbf{County fips code:} Federal Information Processing Standards (FIPS) code for different Counties
    \item \textbf{Age group:} Age groups of patients that include 0 - 17 years, 18 - 49 years, 50 - 64 years, and 65 + years.
    % (see Figure~\ref{fig_age_group}).
    % \begin{figure}[h!]
    %     \centering
    %     \includegraphics[scale = 0.6]{Figures/age_group.png}
    %     \caption{age group attribute distribution.}
    %     \label{fig_age_group}
    % \end{figure}
    \item \textbf{Gender:} Female, Male, Other, Unknown.
    % (see Figure~\ref{fig_Gender}).
    % \begin{figure}[h!]
    %     \centering
    %     \includegraphics[scale = 0.6]{Figures/gender.png}
    %     \caption{Gender attribute distribution.}
    %     \label{fig_Gender}
    % \end{figure}
    \item \textbf{Race:} American Indian/Alaska Native, Asian, Black, Multiple/Other, Native Hawaiian/Other Pacific Islander, White, Unknown (see Table~\ref{tbl_race_attribute} for the distribution of values for race attribute).
    
\begin{table}[h!]
    \centering
    \begin{tabular}{lc}
    \hline
         Race & Count \\
         \hline\hline
         American Indian/ Alaska Native & 94 \\
         Asian & 3067 \\
         Black & 8806 \\
         Multiple/Other & 1833 \\
         Native Hawaiian/Other Pacific Islander & 859 \\
         Unknown & 3081 \\
         White & 78244\\
\hline

    \end{tabular}
    \caption{Race attribute distribution.}
    \label{tbl_race_attribute}
\end{table}
    % (see Figure~\ref{fig_Race}).
    % \begin{figure}[h!]
    %     \centering
    %     \includegraphics[scale = 0.6]{Figures/race.png}
    %     \caption{Race attribute distribution.}
    %     \label{fig_Race}
    % \end{figure}
    \item \textbf{Ethnicity:} Hispanic, Non-Hispanic, Unknown.
    % (see Figure~\ref{fig_Ethnicity}).
    % \begin{figure}[h!]
    %     \centering
    %     \includegraphics[scale = 0.6]{Figures/ethnicity.png}
    %     \caption{Ethnicity attribute distribution.}
    %     \label{fig_Ethnicity}
    % \end{figure}
    \item \textbf{Case positive specimen interval:} Weeks between earliest date and date of first positive specimen collection.
    \item \textbf{Case onset interval:} Weeks between earliest date and date of symptom onset.
    \item \textbf{Process:} Under what process was the case first identified, e,g, Clinical evaluation, Routine surveillance, Contact tracing of case patient, Multiple, Other, Unknown.
    (see Table~\ref{tbl_process_attribute}).
    \begin{table}[h!]
    \centering
    \begin{tabular}{lc}
    \hline
         Process & Count \\
         \hline\hline
         Clinical evaluation & 43768 \\
         Contact tracing of case patient & 6813 \\
         Laboratory reported & 11848 \\
         Multiple & 22595 \\
         Other & 556 \\
         Other detection method (specify) & 164 \\
         Provider reported & 212 \\
         Routine physical examination & 22 \\
         Routine surveillance & 8641 \\
         Unknown & 1365 \\
         \hline

    \end{tabular}
    \caption{Process attribute distribution.}
    \label{tbl_process_attribute}
\end{table}
    % \begin{figure}[h!]
    %     \centering
    %     \includegraphics[scale = 0.6]{Figures/process.png}
    %     \caption{Process attribute distribution.}
    %     \label{fig_process}
    % \end{figure}
    \item \textbf{Exposure:} In the $14$ days prior to illness onset, did the patient have any of the following known exposures e.g. domestic travel, international travel, cruise ship or vessel travel as a passenger or crew member, workplace, airport/airplane, adult congregate living facility (nursing, assisted living, or long-term care facility), school/university/childcare center, correctional facility, community event/mass gathering, animal with confirmed or suspected COVID-19, other exposure, contact with a known COVID-19 case? 
    Possible values for this attribute are  Yes and Unknown.
    % (see Figure~\ref{fig_exposure}).
    % \begin{figure}[h!]
    %     \centering
    %     \includegraphics{Figures/exposure.png}
    %     \caption{Exposure attribute distribution.}
    %     \label{fig_exposure}
    % \end{figure}
    \item \textbf{Current status:} What is the current status of this person? Possible values are Laboratory-confirmed case, Probable case.
    % (see Figure~\ref{fig_hospital}).
    % \begin{figure}[h!]
    %     \centering
    %     \includegraphics{Figures/status_lab.png}
    %     \caption{Current Status attribute distribution.}
    %     \label{fig_hospital}
    % \end{figure}
    \item \textbf{Symptom status:} What is the symptom status of this person? Possible values are Asymptomatic, Symptomatic, Unknown, Missing.
    
    \item \textbf{Hospital:} Was the patient hospitalized? Possible values are Yes, No, Unknown.
    % (see Figure~\ref{fig_hospital}). 

    % \begin{figure}[h!]
    %     \centering
    %     \includegraphics{Figures/hospital.png}
    %     \caption{Hospital attribute distribution.}
    %     \label{fig_hospital}
    % \end{figure}
    \item \textbf{ICU:} Was the patient admitted to an intensive care unit (ICU)? Possible values are Yes, No, Unknown.
    % (see Figure~\ref{fig_icu}). 

    % \begin{figure}[h!]
    %     \centering
    %     \includegraphics{Figures/icu.png}
    %     \caption{ICU attribute distribution.}
    %     \label{fig_icu}
    % \end{figure}
    \item \textbf{Death/Deceased:} This attribute highlights whether the patient die as a result of this illness. The possible values are ``Yes", ``No", and ``Unknown".
    % (see Figure~\ref{fig_death}). 

    % \begin{figure}[h!]
    %     \centering
    %     \includegraphics{Figures/death.png}
    %     \caption{Death attribute distribution.}
    %     \label{fig_death}
    % \end{figure}
    \item \textbf{Underlying Conditions:} This attribute highlights if the patient have single or multiple medical conditions and risk behaviors. These conditions includes diabetes mellitus, hypertension, severe obesity (occurs when BMI is greater than $40$), cardiovascular disease, chronic renal disease, chronic liver disease, chronic lung disease, other chronic diseases, immunosuppressive condition, autoimmune condition, current smoker, former smoker, substance abuse or misuse, disability, psychological/psychiatric, pregnancy, other. The possible values are for this attribute are ``Yes" and ``No".
    % (see Figure~\ref{fig_Underlying_Conditions}). 

    % \begin{figure}[h!]
    %     \centering
    %     \includegraphics{Figures/underlying_condition.png}
    %     \caption{Underlying Conditions distribution.}
    %     \label{fig_Underlying_Conditions}
    % \end{figure}

\end{enumerate}
The Distributions of values for different attributes is shown in Figure~\ref{fig_all_pie_charts}.
To check if there is any natural clustering in Clinical Data1, we use t-distributed stochastic neighbor
embedding (t-SNE) approach~\cite{van2008visualizing}. We maps input data to 2d real vectors representation using t-SNE and Deceased attribute (for Clinical Data1) as class label (see Figure~\ref{fig_t_sne}). We can observe in the figure that there is no visible clustering corresponding to different values of the deceased attribute. All values (No, Yes, and Unknown) are scattered around in the whole plot. This behavior shows that performing any ML task on such data will not directly give us efficient results (since the data is not properly grouped together).

% plot (see Figure~\ref{fig_t_sne}) with using Deceased attribute as class label.
% \todo{Murray: should we comment more on this t-SNE plot, i.e., we do or do not see a natural clustering?}

\subsection{Clinical Data2}
We got the Clinical Data2 data from~\cite{alakus2020comparison}. 
This study used a laboratory dataset of patients with COVID-19 in the Israelita Albert Einstein Hospital in Sao Paulo, Brazil. The patient samples were collected to identify who were infected by COVID-19 in the beginning of year $2020$. The laboratory dataset contains information on $608$ patients with $18$ laboratory findings. In this dataset, $520$ had no findings, and $80$ were patients with COVID-19. The attributes are Red blood Cells, Hemoglobin, Platelets, Hematocrit, Aspartate transaminase, Lymphocytes, Monocytes, Sodium, Urea, Basophils, Creatinine, Serum Glucose, Alanine transaminase, Leukocytes, Potassium, Eosinophils, Proteina C reativa mg/dL, Neutrophils, SARS-Cov-2 exam result (positive or negative). All the attributes (other than ``SARS-Cov-2 exam result") contains integer values. 

\subsection{Evaluation Metrics}
To measure the performance of underlying machine learning classifiers, we use different evaluation metrics such as Average Accuracy, Precision, Recall, weighted and Macro F1, and Receiver Operator Curve (ROC) Area Under the Curve (AUC). We also computed the training runtime of all ML models to see which model is the best in terms of runtime.

\section{Results and Discussion}\label{sec_results}
The results for Clinical Data1 are given in Table~\ref{tbl_clinical_data_1}.
For classifying the Deceased attribute, we can see that all methods are able to classify the label (Deceased attribute) with very high accuracy (accuracy $>90$ in most of the cases). Note that feature selection based models are not only better in terms of prediction accuracy, but also outperforms the setting in which we are not using any feature selection approach (No Feat. Selec.). Also, Boruta feature selection model is outperforming all other feature selection approaches. In terms of training runtime, RFF with Logistic Regression classifier is performing better than the other classifiers.

\begin{table}[h!]
    \centering
    \begin{tabular}{ccp{0.5cm}p{0.5cm}p{0.5cm}p{1.1cm}p{1.1cm}p{0.5cm}|p{1.1cm}}
    \hline
        & & Acc. & Prec. & Recall & F1 (Weighted) & F1 (Macro) & ROC AUC & Train Time (Sec.) \\
        \hline \hline
        % SVM & 0.93 & 0.91 & 0.93 & 0.92 & 0.46 & 0.60 & 1294.57 \\
        \multirow{6}{*}{No Feat. Selec.}
        & NB & 0.78 & 0.93 & 0.78 & 0.83 & 0.49  & 0.80 & 0.19 \\
        & MLP & 0.94 & 0.93 & 0.94 & 0.93 & 0.59 & 0.66 & 35.28 \\
        & KNN & 0.94 & 0.93 & 0.94 & 0.93 & 0.60 & 0.69 & 4.71 \\
        & RF & 0.94 & 0.94 & 0.94 & 0.94 & 0.64 & 0.71 & 4.88 \\
        & LR & 0.93 & 0.87 & 0.93 & 0.90 & 0.32 & 0.50 & 1.38 \\
        & DT & 0.93 & 0.93 & 0.93 & 0.93 & 0.62 & 0.73 & 0.37 \\ 
        \hline
        \multirow{6}{*}{Boruta} & 
        NB & 0.83 & 0.94 & 0.83 & 0.87 & 0.54 & \textbf{0.81} & 0.149 \\
        & MLP & 0.94 & 0.93 & 0.94 & 0.93 & 0.58 & 0.66 & 22.76 \\
        & KNN & 0.94 & 0.94 & 0.94 & 0.94 & 0.62 & 0.70 & 1.814 \\
        & RF & \textbf{0.95} & \textbf{0.94} & \textbf{0.95} & \textbf{0.94} & \textbf{0.64} & 0.72 & 3.346 \\
        & LR & 0.93 & 0.89 & 0.93 & 0.90 & 0.33 & 0.50 & 0.968 \\
        & DT & 0.94 & 0.94 & 0.94 & 0.94 & 0.64 & 0.73 & 0.227 \\
        \hline
        \multirow{6}{*}{RR}
        & NB & 0.84 & 0.93 & 0.84 & 0.87 & 0.45 & 0.72 & \textbf{0.129} \\
        & MLP & 0.93 & 0.87 & 0.93 & 0.90 & 0.32 & 0.50 & 5.658 \\
        & KNN & 0.93 & 0.92 & 0.93 & 0.92 & 0.48 & 0.60 & 1.660 \\
        & RF & 0.94 & 0.93 & 0.94 & 0.93 & 0.51 & 0.64 & 2.214 \\
        & LR & 0.93 & 0.87 & 0.93 & 0.90 & 0.32 & 0.50 & 0.338 \\
        & DT & 0.94 & 0.93 & 0.94 & 0.93 & 0.51 & 0.64 & 0.154 \\
        \hline
        \multirow{6}{*}{RFF}
        & NB & 0.93 & 0.87 & 0.93 & 0.90 & 0.32 & 0.50 & 0.144 \\
        & MLP & 0.93 & 0.89 & 0.93 & 0.90 & 0.32 & 0.50 & 24.22 \\
        & KNN & 0.93 & 0.91 & 0.93 & 0.92 & 0.45 & 0.58 & 3.280 \\
        & RF & 0.94 & 0.93 & 0.94 & 0.93 & 0.56 & 0.64 & 27.87 \\
        & LR & 0.93 & 0.87 & 0.93 & 0.90 & 0.32 & 0.50 & 0.261 \\
        & DT & 0.91 & 0.92 & 0.91 & 0.91 & 0.51 & 0.65 & 1.461 \\
        \hline
        \multirow{1}{*}{Keras Class.}
        & - & 0.93 & 0.87 & 0.93 & 0.90 & 0.32 & 0.50 & 11.582 \\

        \hline
    \end{tabular}
    \caption{Classification Results for Clinical Data1. Best values are shown in bold. In terms of training time, each classifier's runtime is compared separately and best values for each of them are shown in bold.}
    \label{tbl_clinical_data_1}
\end{table}

The results for Clinical Data2 are given in Table~\ref{tbl_clinical_data_2}. For classifying whether a patient is COVID-19 positive or negative, we can see that random forest classifier with Boruta feature selection approach outperforms all other feature selection methods and also the deep learning model. In terms of runtime, logistic regression classifier with RFF is outperforming other approaches.

% In case of both datasets, we observe that Boruta is able to preserve more useful information in the data as compared with RR and RFF.

\begin{remark}
  We note that the deep learning model is slightly worse that the traditional classifiers in case of Clinical Data1 while performing worst on Clinical Data2. This is because the deep learning models are usually ``Data Hungry" and require a lot more data to efficiently learn the patters in the data. Since we have small number of data points in both datasets, the deep learning model is unable to beat the traditional classification algorithms.
\end{remark}

\begin{table}[h!]
    \centering
    \begin{tabular}{ccp{0.5cm}p{0.5cm}p{0.5cm}p{1.1cm}p{1.1cm}p{0.5cm}|p{1.1cm}}
    \hline
        & & Acc. & Prec. & Recall & F1 (Weighted) & F1 (Macro) & ROC AUC & Train Time (Sec.) \\
        \hline \hline
        % SVM & 0.93 & 0.91 & 0.93 & 0.92 & 0.46 & 0.60 & 1294.57 \\
        \multirow{6}{*}{No Feat. Selec.}& 
        % SVM & 0.84 & 0.81 & 0.84 & 0.81 & 0.61 & 0.59 & 0.010 \\
        NB & 0.89 & 0.88 & 0.89 & 0.88 & 0.71 & 0.70 & 0.025 \\
        & MLP & 0.86 & 0.85 & 0.86 & 0.85 & 0.65 & 0.64 & 1.327 \\
        & KNN & 0.88 & 0.87 & 0.88 & 0.87 & 0.68 & 0.66 & 0.013 \\
        & RF & 0.85 & 0.79 & 0.85 & 0.82 & 0.49 & 0.50 & 0.178 \\
        & LR & 0.87 & 0.86 & 0.87 & 0.86 & 0.65 & 0.63 & 0.013 \\
        & DT & 0.81 & 0.81 & 0.81 & 0.81 & 0.56 & 0.56 & 0.01 \\

        \hline
        \multirow{6}{*}{Boruta} & 
        % SVM & 0.89 & 0.86 & 0.89 & 0.86 & 0.65 & 0.62 \\
        NB & 0.83 & 0.89 & 0.83 & 0.85 & 0.71 & 0.79 & 0.01 \\
        & MLP & 0.87 & 0.89 & 0.87 & 0.88 & 0.75 & 0.78 & 1.621 \\
        & KNN & 0.86 & 0.85 & 0.86 & 0.86 & 0.66 & 0.66 & 0.015 \\
        & RF & \textbf{0.91} & \textbf{0.90} & \textbf{0.91} & \textbf{0.90} & \textbf{0.77} & \textbf{0.74} & 0.125 \\
        & LR & 0.87 & 0.88 & 0.87 & 0.88 & 0.73 & 0.74 & 0.01 \\
        & DT & 0.85 & 0.86 & 0.85 & 0.86 & 0.68 & 0.69 & 0.007 \\

        \hline
        \multirow{6}{*}{RR} &
        NB & 0.83 & 0.80 & 0.83 & 0.81 & 0.57 & 0.56 & 0.016 \\
        & MLP & 0.85 & 0.84 & 0.85 & 0.85 & 0.67 & 0.66 & 1.024 \\
        & KNN & 0.85 & 0.84 & 0.85 & 0.84 & 0.66 & 0.64 & 0.01 \\
        & RF & 0.85 & 0.83 & 0.85 & 0.84 & 0.65 & 0.64 & 0.137 \\
        & LR & 0.87 & 0.84 & 0.87 & 0.84 & 0.61 & 0.59 & 0.009 \\
        & DT & 0.82 & 0.82 & 0.82 & 0.82 & 0.62 & 0.62 & 0.009 \\

        \hline
        \multirow{6}{*}{RFF} & 
        NB & 0.89 & 0.78 & 0.89 & 0.83 & 0.47 & 0.50 & 0.022 \\
        & MLP & 0.77 & 0.79 & 0.77 & 0.78 & 0.48 & 0.48 & 1.565 \\
        & KNN & 0.86 & 0.80 & 0.86 & 0.83 & 0.50 & 0.51 & 0.019 \\
        & RF & 0.88 & 0.78 & 0.88 & 0.83 & 0.47 & 0.50 & 0.163 \\
        & LR & 0.89 & 0.78 & 0.89 & 0.83 & 0.47 & 0.50 & \textbf{0.008} \\
        & DT & 0.73 & 0.80 & 0.73 & 0.76 & 0.50 & 0.52 & 0.009 \\

        \hline
        \multirow{1}{*}{Keras Class.} & - &
        0.83 & 0.76 & 0.83 & 0.79 & 0.48 & 0.50 & 10.928 \\

        \hline
    \end{tabular}
    \caption{Classification Results for Clinical Data2. Best values are shown in bold. In terms of training time, each classifier's runtime is compared separately and best values for each of them are shown in bold.}
    \label{tbl_clinical_data_2}
\end{table}

\section{Importance of Attributes}\label{sec_attribute_importance}
\begin{figure}[h!]
  \centering
  % \begin{subfigure}{.33\textwidth}
  \centering
  \includegraphics{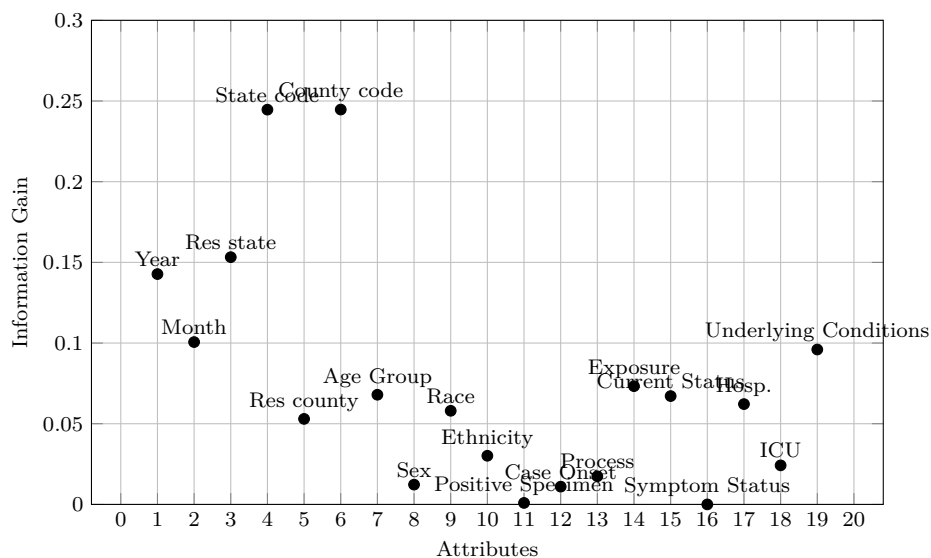}
  \caption{Information Gain of different attributes with respect to Class label (deceased attribute) for Clinical Data1.}
  \label{fig_attribute_correlation}
\end{figure}

To evaluate importance positions in spike sequences, we find the importance of each attribute with respect to class label (for Clinical Data1). For this purpose, we computed the Information Gain (IG) between each attribute and the true class label. 
The IG is defined as follows:
\begin{equation}
    IG(Class,position) = H(Class) - H(Class | position)
\end{equation}
\begin{equation}
    H = \sum_{ i \in Class} -p_i \log p_i
\end{equation}
where $H$ is the entropy, and $p_i$ is the probability of the class $i$. The IG values for each attribute is given in Figure~\ref{fig_attribute_correlation}.

What is particularly interesting is that State and County code of the
are two major predictors of patient outcome.  This is likely due to
the current vaccination situation in the US, which varies quite widely
from county to county~\cite{nyt_url}.

\section{Conclusion}
%----------------------------------------------------------------------
\label{sec_conclusion}

We propose an efficient model for the classification of COVID-19
patients using efficient feature selection methods and machine
learning classification algorithms. We show that with Boruta for
feature selection, the simple classification algorithms like random
forest can also beat the deep learning model when dataset size is not
too big. We also show the importance of each attribute in the Clinical
Data1 by computing the information gain values for each attribute
corresponding to class label. In the future, we will extract more data
and apply other sophisticated deep learning models such as LSTM and
GRU to improve the predictive performance. We will also use other
factors such as weather along with the clinical data to further
improve the classification results.

These results have many practical meanings.  The most direct
real-world application of the machine learning model is to provide
support to medical doctors during the COVID-19 pandemic.  By
predicting the risk level of individual patients, our model enables
clinicians to wisely assign, in real-time, limited medical resources,
especially during periods of medical shortage, and provide immediate
treatment to the most vulnerable groups.  With the help of the risk
prediction system, clinicians learn which individual patients may be
in danger of death and can thus conduct personalized prevention
treatment in due time.  Moreover, our research can be used to build a
general clinical decision support system that serves not only COVID-19
but also other potential future pandemics.  The patterns found in this
data may also help biologies to design effective vaccines or
vaccination strategies.
%Finally, our study provides an approach to scale the machine learning
%model to datasets of a larger size, and a feature selection method to
%reduce the runtime of.
Finally, these methodologies can be applied for future studies on big
data and machine learning in the broader sense.


\bibliographystyle{spmpsci}
\bibliography{death_classification}

\begin{thebibliography}{10}
\providecommand{\url}[1]{{#1}}
\providecommand{\urlprefix}{URL }
\expandafter\ifx\csname urlstyle\endcsname\relax
  \providecommand{\doi}[1]{DOI~\discretionary{}{}{}#1}\else
  \providecommand{\doi}{DOI~\discretionary{}{}{}\begingroup
  \urlstyle{rm}\Url}\fi

\bibitem{abdulkareem2021realizing}
Abdulkareem, K.H., Mohammed, M.A., Salim, A., Arif, M., Geman, O., Gupta, D.,
  Khanna, A.: Realizing an effective covid-19 diagnosis system based on machine
  learning and iot in smart hospital environment.
\newblock IEEE Internet of Things Journal  (2021)

\bibitem{AHMAD2020Combinatorial}
Ahmad, M., Ali, S., Tariq, J., Khan, I., Shabbir, M., Zaman, A.: Combinatorial
  trace method for network immunization.
\newblock Information Sciences \textbf{519}, 215--228 (2020)

\bibitem{alakus2020comparison}
Alakus, T.B., Turkoglu, I.: Comparison of deep learning approaches to predict
  covid-19 infection.
\newblock Chaos, Solitons \& Fractals \textbf{140}, 110120 (2020)

\bibitem{albahri2020role}
Albahri, A.S., Hamid, R.A., Alwan, J.K., Al-Qays, Z., Zaidan, A., Zaidan, B.,
  Albahri, A., AlAmoodi, A., Khlaf, J.M., Almahdi, E., et~al.: Role of
  biological data mining and machine learning techniques in detecting and
  diagnosing the novel coronavirus (covid-19): a systematic review.
\newblock Journal of medical systems \textbf{44}, 1--11 (2020)

\bibitem{ali2021classifying}
Ali, S., Bello, B., Patterson, M.: Classifying covid-19 spike sequences from
  geographic location using deep learning.
\newblock arXiv preprint arXiv:2110.00809  (2021)

\bibitem{ali2021effective}
Ali, S., Khan, M.A., Khan, I., Patterson, M., et~al.: Effective and scalable
  clustering of {SARS-CoV-2} sequences.
\newblock To appear at: International Conference on Big Data Research (ICBDR)
  (2021)

\bibitem{ali2019short}
Ali, S., Mansoor, H., Arshad, N., Khan, I.: Short term load forecasting using
  smart meter data.
\newblock In: International Conference on Future Energy Systems (e-Energy), pp.
  419--421 (2019)

\bibitem{Ali2020ShortTerm}
Ali, S., Mansoor, H., Khan, I., Arshad, N., Khan, M., Faizullah, S.: Short-term
  load forecasting using {AMI} data.
\newblock CoRR \textbf{abs/1912.12479} (2020)

\bibitem{ali2021spike2vec}
Ali, S., Patterson, M.: Spike2vec: An efficient and scalable embedding approach
  for covid-19 spike sequences.
\newblock CoRR \textbf{arXiv:2109.05019} (2021)

\bibitem{ali2021k}
Ali, S., Sahoo, B., Ullah, N., Zelikovskiy, A., Patterson, M., Khan, I.: A
  k-mer based approach for {SARS-Cov-2} variant identification.
\newblock To Appear at: International Symposium on Bioinformatics Research and
  Applications (ISBRA)  (2021)

\bibitem{ali2021predicting}
Ali, S., Shakeel, M., Khan, I., Faizullah, S., Khan, M.: Predicting attributes
  of nodes using network structure.
\newblock ACM Transactions on Intelligent Systems and Technology (TIST)
  \textbf{12}(2), 1--23 (2021)

\bibitem{validationSetApproach}
Devijver, P., Kittler, J.: Pattern recognition: A statistical approach.
\newblock In: London, GB: Prentice-Hall, pp. 1--448 (1982)

\bibitem{fung2021predictive}
Fung, D.L., Hoi, C.S., Leung, C.K., Zhang, C.Y.: Predictive analytics of
  covid-19 with neural networks.
\newblock In: 2021 International Joint Conference on Neural Networks (IJCNN),
  pp. 1--8 (2021)

\bibitem{gisaid_website_url}
{GISAID Website}: .
\newblock \url{https://www.gisaid.org/}  (Accessed: 10-10-2021)

\bibitem{grover2016node2vec}
Grover, A., Leskovec, J.: node2vec: Scalable feature learning for networks.
\newblock In: International Conference on Knowledge Discovery \& Data Mining
  (KDD), pp. 855--864 (2016)

\bibitem{hassan2021locally}
Hassan, I.U., Haseeb, A., Ali, S.: Locally weighted mean phase angle (lwmpa)
  based tone mapping quality index (tmqi-3).
\newblock Accepted at: International Conference on Intelligent Vision and
  Computing (ICIVC)  (2021)

\bibitem{hoerl1975ridge}
Hoerl, A.E., Kannard, R.W., Baldwin, K.F.: Ridge regression: some simulations.
\newblock Communications in Statistics-Theory and Methods \textbf{4}(2),
  105--123 (1975)

\bibitem{kursa2010feature}
Kursa, M.B., Rudnicki, W.R., et~al.: Feature selection with the boruta package.
\newblock J Stat Softw \textbf{36}(11), 1--13 (2010)

\bibitem{kuzmin2020machine}
Kuzmin, K., et~al.: Machine learning methods accurately predict host
  specificity of coronaviruses based on spike sequences alone.
\newblock Biochemical and Biophysical Research Communications \textbf{553}(3),
  553--558 (2020)

\bibitem{leung2020machine}
Leung, C.K., Chen, Y., Hoi, C.S., Shang, S., Cuzzocrea, A.: Machine learning
  and olap on big covid-19 data.
\newblock In: 2020 IEEE International Conference on Big Data (Big Data), pp.
  5118--5127 (2020)

\bibitem{leung2020big}
Leung, C.K., Chen, Y., Shang, S., Deng, D.: Big data science on covid-19 data.
\newblock In: 2020 IEEE 14th International Conference on Big Data Science and
  Engineering (BigDataSE), pp. 14--21 (2020)

\bibitem{leung2020data}
Leung, C.K., Fung, D.L., Mushtaq, S.B., Leduchowski, O.T., Bouchard, R.L., Jin,
  H., Cuzzocrea, A., Zhang, C.Y.: Data science for healthcare predictive
  analytics.
\newblock In: Proceedings of the 24th Symposium on International Database
  Engineering \& Applications, pp. 1--10 (2020)

\bibitem{li2020using}
Li, W.T., Ma, J., Shende, N., Castaneda, G., Chakladar, J., Tsai, J.C.,
  Apostol, L., Honda, C.O., Xu, J., Wong, L.M., et~al.: Using machine learning
  of clinical data to diagnose covid-19: a systematic review and meta-analysis.
\newblock BMC medical informatics and decision making \textbf{20}(1), 1--13
  (2020)

\bibitem{van2008visualizing}
Van~der M., L., Hinton, G.: Visualizing data using t-{SNE}.
\newblock Journal of Machine Learning Research (JMLR) \textbf{9}(11) (2008)

\bibitem{mcdonald2009ridge}
McDonald, G.C.: Ridge regression.
\newblock Wiley Interdisciplinary Reviews: Computational Statistics
  \textbf{1}(1), 93--100 (2009)

\bibitem{melnyk2021alpha}
Melnyk, A., Mohebbi, F., Knyazev, S., Sahoo, B., Hosseini, R., Skums, P.,
  Zelikovskiy, A., Patterson, M.D.: From alpha to zeta: Identifying variants
  and subtypes of sars-cov-2 via clustering.
\newblock bioRxiv  (2021)

\bibitem{nyt_url}
{NewYork Times (NYT)}:
  \url{https://www.nytimes.com/interactive/2020/us/covid-19-vaccine-doses.html}
  (2021).
\newblock [Online; Accessed: 15-10-2021]

\bibitem{rahimi2007random}
Rahimi, A., Recht, B., et~al.: Random features for large-scale kernel machines.
\newblock In: NIPS, vol.~3, p.~5 (2007)

\bibitem{Shakeel2020LanguageIndependent}
Shakeel, M.H., Faizullah, S., Alghamidi, T., Khan, I.: Language independent
  sentiment analysis.
\newblock In: 2019 International Conference on Advances in the Emerging
  Computing Technologies (AECT), pp. 1--5 (2020)

\bibitem{Shakeel2019MultiBilingual}
Shakeel, M.H., Karim, A., Khan, I.: A multi-cascaded deep model for bilingual
  sms classification.
\newblock In: International Conference on Neural Information Processing, pp.
  287--298 (2019)

\bibitem{Shakeel2020Multi}
Shakeel, M.H., Karim, A., Khan, I.: A multi-cascaded model with data
  augmentation for enhanced paraphrase detection in short texts.
\newblock Information Processing \& Management \textbf{57}(3), 102204 (2020)

\bibitem{ullah2020effect}
Ullah, A., Ali, S., Khan, I., Khan, M., Faizullah, S.: Effect of analysis
  window and feature selection on classification of hand movements using {EMG}
  signal.
\newblock In: SAI Intelligent Systems Conference (IntelliSys), pp. 400--415
  (2020)

\bibitem{yang2018multi}
Yang, L., Guo, Y., Cao, X.: Multi-facet network embedding: Beyond the general
  solution of detection and representation.
\newblock In: AAAI Conference on Artificial Intelligence (AAAI), pp. 499--506
  (2018)

\end{thebibliography}

% \begin{wrapfigure}{l}{60mm} 
%     \includegraphics[scale = 0.6]{Figures/sarwan.png}
%   \end{wrapfigure}\par
%   \textbf{Author A} is a well-known author in the field of the journal scope. His/Her research interests include interest 1, interest 2.\par

\begin{figure}[H]
  \begin{minipage}[c]{0.3\textwidth}
    \includegraphics[scale = 0.4]{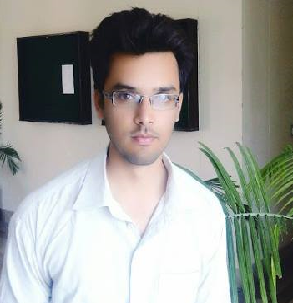}
  \end{minipage}\hfill
  \begin{minipage}[c]{0.67\textwidth}
    \caption*{
		\textbf{Sarwan Ali} is a Ph.D student at Georgia State University working in the field of Bioinformatics, Data mining, Big data, and Machine Learning. He has published many articles in reputed journals and conferences in these domains. In free time, he likes to watch movies.
    } \label{fig:03-03}
  \end{minipage}
\end{figure}

\begin{figure}[H]
  \begin{minipage}[c]{0.3\textwidth}
    \includegraphics[scale = 0.22]{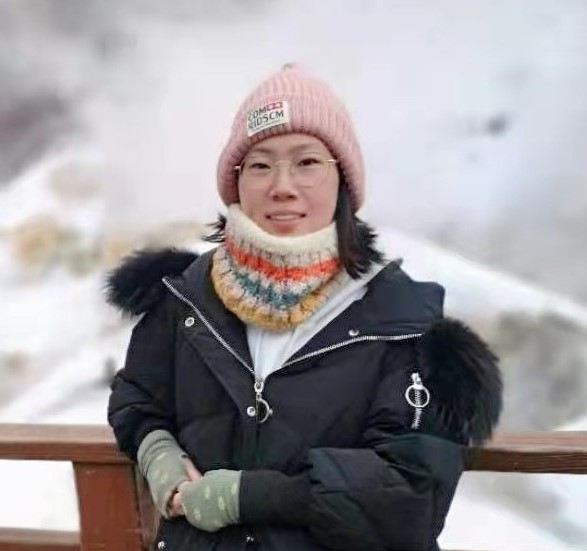}
  \end{minipage}\hfill
  \begin{minipage}[c]{0.67\textwidth}
    \caption*{
		\textbf{Yijing Zhou} is an Undergraduate student at Georgia State University, studying Computer Science and Biology. She is working in the fields of Bioinformatics, Computational Biology, and Combinatorics.
    } \label{fig:03-03}
  \end{minipage}
\end{figure}

\begin{figure}[H]
  \begin{minipage}[c]{0.3\textwidth}
    \includegraphics[scale = 0.3]{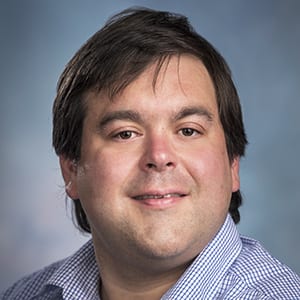}
  \end{minipage}\hfill
  \begin{minipage}[c]{0.67\textwidth}
    \caption*{
		\textbf{Murray Patterson} is an Assistant Professor at Georgia State University working in the fields of Bioinformatics, Computational Biology, Algorithms, and Combinatorics.
    } \label{fig:03-03}
  \end{minipage}
\end{figure}

\end{document}